\documentclass[10pt,twocolumn,letterpaper]{article}

 \usepackage[pagenumbers]{cvpr} % To force page numbers, e.g. for an arXiv version
\usepackage{epsfig}
\usepackage{graphicx}
\usepackage{amsmath}
\usepackage{amssymb}
\usepackage{booktabs}
\usepackage{multirow}
\usepackage{multicol}
\usepackage{comment}
\usepackage{caption}
\usepackage{subcaption}
\usepackage[dvipsnames]{xcolor}
\usepackage[pagebackref,breaklinks,colorlinks]{hyperref}

\usepackage[dvipsnames]{xcolor}

\definecolor{cvprblue}{rgb}{0.21,0.49,0.74}

\def\paperID{*****} % *** Enter the Paper ID here
\def\confName{CVPR}
\def\confYear{2024}

\title{ICSVR: Investigating Compositional and Syntactic Understanding in Video Retrieval Models}

\author{
Avinash Madasu
\qquad
Vasudev Lal\\
Cognitive AI, Intel Labs\\ {\tt\small \{avinash.madasu, vasudev.lal\}@intel.com}
}

\begin{document}
\maketitle
\begin{abstract}
Video retrieval (VR) involves retrieving the ground truth video from the video database given a text caption or vice-versa. The two important components of compositionality: objects \& attributes and actions are joined using correct syntax to form a proper text query. These components (objects \& attributes, actions and syntax)  each play an important role to help distinguish among videos and retrieve the correct ground truth video. However, it is unclear what is the effect of these components on the video retrieval performance.  We therefore, conduct a systematic study to evaluate the compositional and syntactic understanding of video retrieval models on standard  benchmarks such as MSRVTT, MSVD and DIDEMO.  The study is performed on two categories of video retrieval models: (i) which are pre-trained on video-text pairs and fine-tuned on downstream video retrieval datasets (Eg. Frozen-in-Time, Violet, MCQ etc.) (ii) which adapt pre-trained image-text representations like CLIP for video retrieval (Eg. CLIP4Clip, XCLIP, CLIP2Video etc.). Our experiments reveal that actions and syntax play a minor role compared to objects \& attributes in video understanding. Moreover, video retrieval models that use pre-trained image-text representations (CLIP) have better syntactic and compositional understanding as compared to models pre-trained on video-text data. The code is available at \url{https://github.com/IntelLabs/multimodal_cognitive_ai/tree/main/ICSVR}.
\end{abstract}    
\section{Introduction}
\label{sec:intro}
Video-retrieval (VR) is the task of retrieving videos for a given text caption or given a video, retrieve the corresponding text caption. This involves understanding important details such as \textbf{\textcolor{Salmon}{objects \& attributes}} (Eg: two women and a man, red shirt guy), \textbf{\textcolor{LimeGreen}{actions}} (Eg. playing, standing, talking etc.)  in the text caption and the video. In vision it is referred to as compositional reasoning \cite{lillo2014discriminative,ji2020action,grunde2021agqa,li2022compositional,chencomphy}, i.e. representing the image or video requires the understanding of primitive concepts that make them. In the recent years, new benchmarks \cite{johnson2017clevr,hudson2019gqa,chen2020cops,moisio2023evaluating} have been proposed to measure the compositional capabilities of foundational image models. The compositionality in these models is measured by creating new text captions from the original text captions using word ordering \cite{thrush2022winoground}, word substitutions \cite{parcalabescu2022valse}, negative pairs \cite{hendricks2021probing}, image-text mismatch \cite{shekhar2017foil}.

When compared to images, measuring compositionality is a lot harder in videos. There are multiple reasons to this: First,  videos are made-up of time-series image frames with multiple \textbf{\textcolor{Salmon}{objects \& attributes}} and \textbf{\textcolor{LimeGreen}{actions}} unlike images. Therefore, methods like creating negative pairs, mismatching pairs etc. used for evaluating compositionality in image-language models have very limited scope.  Second, even though tasks based on video question answering (VQA) \cite{hudson2019gqa,chen2020cops} have been proposed to measure the compositionality, recent studies \cite{huang2018makes,rawal2023revealing,buch2022revisiting,fan2019heterogeneous,gandhi2022measuring} have shown that these datasets exhibit single frame bias. Most of the previous works \cite{thrush2022winoground,parcalabescu2022valse,hendricks2021probing} focus on understanding the compositionality  of image-text models. It mainly involves experimenting with \textbf{\textcolor{Salmon}{objects \& attributes}} in the text captions and retrieving the images. However,  \textbf{\textcolor{LimeGreen}{actions}} play a crucial role in when retrieving videos using text captions. Another important aspect which is often overlooked in the previous studies is the \textbf{\textcolor{SkyBlue}{syntactics}}. For example consider the query 
\textit{``\textbf{\textcolor{SkyBlue}{a}} \textbf{\textcolor{Salmon}{guy}} \textbf{\textcolor{LimeGreen}{wearing}} \textbf{\textcolor{SkyBlue}{a}} \textbf{\textcolor{Salmon}{red shirt}} \textbf{\textcolor{LimeGreen}{drives}} \textbf{\textcolor{SkyBlue}{a}} \textbf{\textcolor{Salmon}{car}} \textbf{\textcolor{SkyBlue}{while}} \textbf{\textcolor{LimeGreen}{talking}}''}, the \textbf{\textcolor{Salmon}{objects \& attributes}} are \textbf{\textcolor{Salmon}{guy}}, \textbf{\textcolor{Salmon}{red shirt}} and \textbf{\textcolor{Salmon}{car}}, the \textbf{\textcolor{LimeGreen}{actions}} are \textbf{\textcolor{LimeGreen}{wearing}}, \textbf{\textcolor{LimeGreen}{driving}} and \textbf{\textcolor{LimeGreen}{talking}} and rest of the words (\textbf{\textcolor{SkyBlue}{a}}, \textbf{\textcolor{SkyBlue}{while}})  form the \textbf{\textcolor{SkyBlue}{syntactics}} of the text captions. The video retrieval models can comprehend such queries because of the accurate \textbf{\textcolor{SkyBlue}{syntactic}} and compositionality (\textbf{\textcolor{Salmon}{objects \& attributes}} and \textbf{\textcolor{LimeGreen}{actions}}).

Now consider the following scenarios of the text captions in which (i) objects \& attributes are missing (\textit{\textbf{\textcolor{SkyBlue}{a}}  \textbf{\textcolor{LimeGreen}{wearing}} \textbf{\textcolor{SkyBlue}{a}} \textbf{\textcolor{LimeGreen}{drives}} \textbf{\textcolor{SkyBlue}{a}} \textbf{\textcolor{SkyBlue}{while}} \textbf{\textcolor{LimeGreen}{talking}}}) (ii) actions are missing (\textit{\textbf{\textcolor{SkyBlue}{a}} \textbf{\textcolor{Salmon}{guy}}  \textbf{\textcolor{SkyBlue}{a}} \textbf{\textcolor{Salmon}{red shirt}}  \textbf{\textcolor{SkyBlue}{a}} \textbf{\textcolor{Salmon}{car}} \textbf{\textcolor{SkyBlue}{while}}}) and (iii) syntactics are missing (\textit{\textbf{\textcolor{Salmon}{guy}} \textbf{\textcolor{LimeGreen}{wearing}}  \textbf{\textcolor{Salmon}{red shirt}} \textbf{\textcolor{LimeGreen}{drives}} \textbf{\textcolor{Salmon}{car}} \textbf{\textcolor{LimeGreen}{talking}}}). This begs an important question: \textbf{What is the effect of each of these scenarios on the video retrieval performance?}

To address this question, we propose a detailed study to evaluate the syntax and compositional understanding of video retrieval models. For this study we create a comprehensive test bed to evaluate the state-of-the-art video retrieval models for compositonality and syntactics. We base this investigation along three axes: Objects \& attributes, actions and syntactics. We propose a set of 10 tasks for these categories: four tasks to evaluate the knowledge of VR models for objects \& attributes (\S \ref{objattr}), three tasks for testing action understanding (\S \ref{action}) and finally, three tasks for syntactic capabilities (\S \ref{syntactics}). Table ~\ref{tab:captions} describes these tasks with an example. We perform a comprehensive evaluation on 12 state-of-the-art video retrieval models belonging to two categories (\S\ref{models}): The first category of models such as Frozen-in-Time (FiT) \cite{bain2021frozen}, MCQ \cite{ge2022bridging} etc. are pre-trained on large scale video datasets and fine-tuned for video retrieval. The second category uses pretrained image features like CLIP for video retrieval namely CLIP4Clip \cite{luo2022clip4clip}, CLIP2Video \cite{fang2021clip2video} etc. These models are tested on three standard video retrieval benchmarks (\S\ref{datasets}) such as MSRVTT \cite{xu2016msr}, MSVD \cite{chen2011collecting} and DiDeMo \cite{anne2017localizing}. 

Our experiments (\S \ref{basic}) reveal that objects \& attributes are the most crucial to video retrieval followed by actions and syntax. Among video retrieval models, CLIP based models have a better compositional and syntactic understanding when compared with pretrained video models. We further perform detailed studies to fully judge how retrieval models perceive each of the components. We find that (\S\ref{objabl}) video retrieval models have a poor understanding of relationship between objects and its attributes. However, they are extremely sensitive to incorrect object references in the captions. Our studies on action understanding (\S\ref{actionabl}) disclose that models have poor sense of action negation and replacing them with incorrect actions lead to slight decrease in video retrieval performance. Finally, we discover (\S\ref{wordorder}) that models perform significantly better even without the right syntax. In summary, our contributions in this paper are as follows:
\begin{itemize}
    \item Ours is the first work to comprehensively investigate the compositional and syntactic understanding of video retrieval models.
    \item For this study, we propose a set of 10 tasks dealing with different aspects of compositionality and syntax.
           \item We perform this analysis on a broad range of 12 state-of-the-art  models and generalize the findings to the video retrieval task.
    \item We establish that video retrieval models exhibit distinct and contrasting behaviours for interpreting various elements in the text captions. 
\end{itemize}

\section{Related Work}
\subsection{Video retrieval}
In the recent years, there has been a tremendous improvement on the task of video-retrieval. This is mainly due to two reasons (i) with the adaption of transformer based models to vision tasks like image classification \cite{he2022masked,liu2021swin,dosovitskiy2020image} (ii) with the availabilty of large scale video-text datasets like HowTo100M \cite{miech2019howto100m}, WebVid-2M \cite{bain2021frozen} and YT180M \cite{zellers2021merlot}. Frozen-in-Time \cite{bain2021frozen} is a dual-stream transformer model pre-trained on WebVid-2M and Conceptual captions-3M \cite{sharma2018conceptual} datasets and fine-tuned for downstream video retrieval.
A prompt based novel pre-training task \cite{li2021align} is proposed to effectively align visual and text features during large scale video-text pre-training. A new pre-training approach Masked Visual-token Modeling (MVM) \cite{fu2021violet} is presented to better model the temporal dependencies among videos for video-retrieval. To incorporate the rich semantic features of the videos, a novel pretext task Multiple Choice Questions (MCQ) is put forward in which the model is trained to answer questions about the video.

In a parallel direction, image features pre-trained on large amounts of image-text data have been adopted for the task of video retrieval. CLIP4Clip \cite{luo2022clip4clip} is an end-to-end trainable video retrieval model based on CLIP \cite{radford2021learning} architecture in which frame features are extracted using clip image encoder and the temporal modelling is performed using a transformer encoder. A two-stage framework CLIP2Video \cite{fang2021clip2video} is proposed to enhance interaction among video features and video-text features for video-retrieval. Madasu et at. \cite{madasu2022improving} used off-the-shelf multi-lingual data to enhance the performance of video-retrieval. All these video-retrieval models haven't been tested for syntactic and compositional understanding. To the best of our knowledge, ours is the first work to comprehensively explore syntactic and compositional understanding of video retrieval models.

\subsection{Syntactics}
Transformer based language models \cite{devlin2019bert,hedeberta,yang2019xlnet} have achieved state-of-the-art results on most natural language understanding tasks \cite{wang2018glue,wang2019superglue}. Hence, there has been a growing interest to explore the morphological capabilities of these models \cite{zhang2020semantics,rogers2021primer,vulic2020probing,pavlick2022semantic,gari2021let}. Since all the video retrieval models use pre-trained language models for encoding text captions, we build upon those works and investigate their syntactic understanding. 

\subsection{Compositionality}
Although vision-language models pretrained on large amounts of data achieved state-of-the-results there has been a growing interest to understand the working of these models \cite{johnson2017clevr,hudson2019gqa,chen2020cops,moisio2023evaluating,hudson2019gqa,suhr2019corpus}. These works mainly focus on compositional knowledge these models by proposing new benchmarks. 
Winoground \cite{thrush2022winoground} dataset was introduced in which a pair of text captions contain the same set of words but pertain to different images. The models are then tested for image and caption match. Another benchmark CREPE \cite{gandhi2022measuring} was put forward to evaluate two aspects of compositionality: systematicity and productivity. This benchmark contains unseen compounds and atoms in the test split to evaluate the models' generalization.
 Parcalabescu et al. \cite{parcalabescu2022valse} proposed VALSE dataset to measure visio-linguistic capabilities of pretrained vision and language models. AGQA-Decomp \cite{gandhi2022measuring} is a new benchmark to measure compositional consistency for the task of Video Question Answering. All these works proposed new benchmarks for compositional reasoning in image-language models. Contrary to these, our work focuses on measuring compositionality of video retrieval models using the standard datasets and doesn't require a new benchmark. Moreover, our experiments are evaluated on 12 models which are significantly higher than the frequency of models used in these works.  
\begin{comment}
Although vision-language models have achieved state-of-the-art results on many multi-modal tasks, several works explored the reason behind their effectiveness. Iki et al. \cite{iki2021effect} analyzed the effect of multimodal representations on natural language understanding (NLU) tasks. A visual winoground dataset \cite{thrush2022winoground} is proposed to understand the visual-linguistic compositionality. Madasu et al. \cite{madasu2023multi} compared the performance of vision supervised language representations to vanilla language representations on several NLU tasks. Yuksekgonul et al. \cite{yuksekgonul2022and} analyzed the compositionality on several image tasks. 
\end{comment}
\section{Compositional and Syntactic Understanding} \label{compsem}
\begin{table*}[]
    \centering
    \begin{tabular}{c| c| c}
          Notation &Caption type & Example\\
    \hline
    $Q$ & Original caption & a guy wearing a red shirt drives a car while talking
 \\
 \hline
     $Q_{objattrrem}$ & Object \& Attribute removal & a \textcolor{red}{wearing a drives a while talking
} \\
$Q_{objshift}$ & Object  shift & a \textcolor{red}{shirt wearing a red car drives a guy while talking
} \\
$Q_{objrep}$ & Object replacement & a \textcolor{red}{surf} wearing a red \textcolor{red}{mars} drives a \textcolor{red}{channel} while talking \\
$Q_{objpartial}$ & Object partial & a \textcolor{red}{wearing a red drives a car while talking
}
 \\
\hline
    $Q_{actrem}$ & Action removal & a guy is \textcolor{red}{a red shirt a car while}
 \\
     $Q_{actneg}$ & Action negation & a guy \textcolor{red}{not} wearing a red shirt \textcolor{red}{not} drives a car while \textcolor{red}{not} talking
 \\
     $Q_{actrep}$ & Action replacement & a guy \textcolor{red}{removing} a red shirt \textcolor{red}{flying} a car while \textcolor{red}{sleeping}
 \\
 \hline
    $Q_{synrem}$ & Syntax removal & \textcolor{red}{guy wearing red shirt drives car talking}
    \\
    $Q_{shuf}$ & Word order shuffle & \textcolor{red}{talking red shirt drives while car a guy a wearing a
} \\
 $Q_{rev}$ & Word order reverse & \textcolor{red}{talking while car a drives shirt red a wearing guy a
} \\
    \end{tabular}
    \caption{Table shows the types of perturbations applied to the text captions. The example text caption is taken from the MSRVTT \cite{xu2016msr} dataset. Red color denotes the change from the original text caption.}
    \label{tab:captions}
\end{table*}
In this section, we first define syntax and compositionality and subsequently establish the evaluation protocol for syntactic and compositional understanding in video retrieval models. For this evaluation, we augment the existing text captions and create new datasets that assess their syntactic and compositional understanding. We explain this protocol using an example test caption ($Q$) \textit{``a guy wearing a red shirt drives a car while talking''} from the MSRVTT dataset. Table ~\ref{tab:captions} summarizes  different augmentation methods used for the proposed study.
\subsection{Compositionality in videos} \label{compo}
A video is composed of multiple objects \& attributes interacting with each other in a similar or different fashion. To retrieve a video, corresponding text caption is passed as an input to the video retrieval model. This text caption typically consists of objects \& attributes and interactions (actions) unique to that particular video. The video retrieval model parses the input caption and computes the matching scores with all the videos. Finally, the video with the highest matching score is the predicted ground truth video. Therefore a video retrieval model should be able to understand each of the objects \& attributes and actions present in the caption. This is called compositionality in the visual world. To evaluate the compositional understanding in video retrieval models, we mainly focus on their ability to parse objects \& attributes and actions. Next we discuss the evaluation protocol to measure compositionality in VR models.
\subsubsection{Object \& Attribute knowledge} \label{objattr}
\textbf{Object \& Attribute removal ($Q_{objattrrem}$):} In this setup, we remove all the objects \& attributes in the original caption $Q$ and the resulting caption is  \textit{``wearing a drives a while talking''}. Here guy, red shirt and car are the objects \& attributes. \\
\textbf{Object shift ($Q_{objshift}$):} To test the VR models ability to relate objects with their attributes, we shift the places of objects in the captions. The modified caption is \textit{``a shirt wearing a red car drives a guy while talking''}.\\
\textbf{Object replacement ($Q_{objrep}$):} We evaluate the VR models sensitivity to objects by randomly replacing the objects with an entirely different objects. The replaced caption is \textit{``a surf wearing a red mars drives a guy channel while talking''}.\\
\textbf{Object partial ($Q_{objpartial}$):} In this setup, the VR models are given access to just 50\% of the objects in the caption. This is to understand if the models perform any shortcuts while retrieving videos. Eg: \textit{``a wearing a red drives a car while talking''}.
Next, we introduce the tasks for evaluating action knowledge in VR models.
\subsubsection{Action knowledge} \label{action}
\textbf{Action removal ($Q_{actrrem}$):} The actions present in the original captions are eliminated. The modified caption is \textit{``a red shirt a car while''} as the actions wearing, drives are removed. This is to understand the influence of actions on the video retrieval performance. \\  
\textbf{Action negation ($Q_{actneg}$):} A negation is added to all the actions in the captions resulting in the new caption  \textit{``a guy \textbf{not} wearing a red shirt \textbf{not} drives a car while \textbf{not} talking''} This tests the VR models ability to comprehend negation in the captions. \\
\textbf{Action replacement ($Q_{actrep}$):} In this setup, the actions are randomly replaced with a different set of actions. The replaced actions are neither antonyms nor synonyms. It checks if the models truly recognize the meaning of the action words.  
Next we present the evaluation protocol for syntactic understanding of VR models. 
\subsection{Syntactic understanding} \label{syntactics}
In the previous section we elucidated the components for compositional reasoning in videos namely objects \& attributes and actions. These components are bind together by syntax there by forming a meaningful caption. Let's consider a part of the example described previously ``a guy wearing a red shirt drives a car'', if the word ``car'' and ``guy'' are interchanged the resulting caption will be  ``a car wearing a red shirt drives a guy'' which is not meaningful. Consequently, syntax also play a crucial role in video retrieval performance along with the compositonality. Subsequently, we put forward the evaluation protocol to measure syntactic understanding in video retrieval models.\\
\textbf{Syntax removal ($Q_{synrem})$:} 
Our first experiment focuses on the effect of syntax on VR models. We modify the caption by keeping just the objects \& attributes, actions and eliminate any meaningful syntax among them. The resulting caption is  \textit{``guy wearing red shirt drives car talking''}. \\
\textbf{Word order shuffle ($Q_{shuf})$:}
In this setup, all the words are shuffled in the caption. This destroys the order of compositionality and syntax. This tests the order sensitivity of VR models. \\
\textbf{Word order reverse ($Q_{rev})$:}
In this setup, we preserve the word order except that in the reverse order. It evaluates the positional knowledge of video retrieval models.
Next, we present the experiment set up for quantifying the compositional and syntactic understanding.
\begin{table*}[ht]
	\begin{center}
		%\resizebox{0.98\textwidth}{!}
		{
		\begin{tabular}{c| c| cccc| cccc} %
	    \hline
        & & \multicolumn{4}{c}{Text-to-Video Retrieval} &\multicolumn{4}{c}{Video-to-Text Retrieval} \\
        \hline
	   Type & Model & Q  & $Q_{actrem}$ & $Q_{objattrrem}$ & $Q_{synrem}$ & Q & $Q_{actrem}$ & $Q_{objattrrem}$ & $Q_{synrem}$ \\
        \hline
	\multirow{5}{*}{Pretrained}    
 & FiT \cite{bain2021frozen} &    26.1 &   22.8 &  5.2 & 20 & 27.9 &  23.7 &  5.8 & 25.7         \\
     & MCQ \cite{ge2022bridging} & 26 &   21.9  &  4.1 & 20.1 & 19.4 &   15.7 &   3.7 & 18.6            \\
   \multirow{3}{*}{video}  & MILES \cite{ge2022miles} &    26 &    21.3 &   3.3 & 19.9 & 17.5   &  15.2 &    2.9 & 17.1          \\
      & VIOLET \cite{fu2021violet} &     35.6  &    29.5 &    0.1  &  25 & -  & - & - & -             \\
      & MVM \cite{fu2023empirical} & 36.3 &   31 &  8.7 &  33.7 & -  & - & - & - \\
	\hline
     \multirow{7}{*}{CLIP \cite{radford2021learning}} & TS2NET \cite{liu2022ts2} &     36  &    30.6 &   6 &  29.3 & 25.4  &    21.2 &   4.3 & 41.4            \\
      & CLIP4Clip \cite{luo2022clip4clip} &     43.4   &    37  & 9.7 & 35.3 & 43.6   &   39  & 10.3 &   39.7            \\
     & CLIP2Video \cite{fang2021clip2video} &     46  & 38.8 &   8.4 & 35.3 & 43  &   38 & 10 & 40.8         \\
      & XCLIP  \cite{ma2022x} &     46.1    & 39.8  &  10.5 &   35.6 & 45.4  &   40.2  &  11 & 42.2            \\
    & XPOOL \cite{gorti2022x} &    46.9 & 39.5 & 7.6 & 36.4 & 44.4  &   39.6 &   11.1 &   42          \\
     & EMCL \cite{jin2022expectation} &    47.8    & 40.8 & 8.2 &  37.4 & 46.2  &   39.5   & 11.6 &     42.8            \\
      & DiCoSA \cite{jin2023text} &     47.9 &   41.3 & 9.1 & 38.3 & 45.9 &   41.2 & 13.4 & 43.1           \\
	   \hline
		\end{tabular}}
		\caption{The table shows the results on MSRVTT \cite{xu2016msr} dataset in both text-to-video and video-to-text retrieval settings. $Q$ denotes the performance (R@1 score) on the original unchanged dataset. $Q_{actrem}$, $Q_{objattrrem}$ and $Q_{synrem}$ is the R@1 score on datasets that have excluded actions, attributes and syntax respectively.}
        \label{tab: MSRVTT_rem}
	\end{center}
\end{table*}
\begin{table*}[ht]
	\begin{center}
		%\resizebox{0.98\textwidth}{!}
		{
		\begin{tabular}{c| c| cccc| cccc} %
	    \hline
       & & \multicolumn{4}{c}{Text-to-Video Retrieval} &\multicolumn{4}{c}{Video-to-Text Retrieval} \\
        \hline
	    Type & Model & Q  & $Q_{actrem}$ & $Q_{objattrrem}$ & $Q_{synrem}$ & Q & $Q_{actrem}$ & $Q_{objattrrem}$ & $Q_{synrem}$ \\
        \hline
        \multirow{5}{*}{Pretrained}
	    & FiT \cite{bain2021frozen} &  36 &  32.7 & 6.9 &  34.9 & 36.1  &   31 & 7.9 &   34.9      \\
     & MCQ \cite{ge2022bridging} & 43.6 &    36.4 & 9 & 42.4 & 40.3 &    33.7 &  9.9 &  39          \\
     \multirow{3}{*}{video} & MILES \cite{ge2022miles} &    44 &   39 &  8.1 & 43.9 & 43.7 &   37.3   & 9.6 &  41.5            \\
     & VIOLET \cite{fu2021violet} &     48.3 & 40.6 & 10.8 &  45.8 & - & - & - &  -            \\
     & MVM \cite{fu2023empirical} & 49.6 &   41.5 & 10.5 &   45.6 & -  & - & -  & - \\
	\hline
    \multirow{5}{*}{CLIP \cite{radford2021learning}}  & TS2NET \cite{liu2022ts2} &     52.8 &   38.5 &  11.8 & 49.4 & 51.2  &  37.1 & 10.7 &  48.6            \\
     & CLIP4Clip \cite{luo2022clip4clip} &     54.5 &  42.1 & 11.9 & 51.9 & 51.8 &   38.5 & 10.9  &  50.7           \\
     & CLIP2Video \cite{fang2021clip2video} &     55.8  &   41.6 & 11.8 &  50.6 & 53.6 &   40.5  &  12.5 & 51.6       \\
    &  XCLIP \cite{ma2022x} &     54  &  39.7 & 12.4 &  49.7 & 54.9 &  42.6 &   13.3 & 48.6          \\
   &  XPOOL \cite{gorti2022x} &    56.1  &   47 & 11.9 &   53.9 & 56.6 &  48  & 12.4 &  53.3 \\
	   \hline
		\end{tabular}}
		\caption{The table shows the results on MSVD \cite{chen2011collecting} dataset in both text-to-video and video-to-text retrieval settings. $Q$ denotes the performance (R@1 score) on the original unchanged dataset. $Q_{actrem}$, $Q_{objattrrem}$ and $Q_{synrem}$ is the R@1 score on datasets that have excluded actions, attributes and syntax respectively.}
        \label{tab: MSVD_rem}
	\end{center}
\end{table*}
\begin{table*}
	\begin{center}
		%\resizebox{0.98\textwidth}{!}
		{
		\begin{tabular}{c| c| cccc| cccc} %
	    \hline
        & & \multicolumn{4}{c}{Text-to-Video Retrieval} &\multicolumn{4}{c}{Video-to-Text Retrieval} \\
        \hline
	     Type & Model & Q  & $Q_{actrem}$ & $Q_{objattrrem}$ & $Q_{synrem}$ & Q & $Q_{actrem}$ & $Q_{objattrrem}$ & $Q_{synrem}$ \\
        \hline
        \multirow{5}{*}{Pretrained} &
	     FiT  \cite{bain2021frozen} &    29.2 &	28 &	4.4 & 27.4 & 28.2	& 27.1 & 5.6 & 27          \\
     & MCQ \cite{ge2022bridging} & 24.6 &  22.3 &   5.6 & 22.8 & 23.8 &   21.2 &  5.2 &  21.4            \\
  \multirow{3}{*}{video}  &  MILES  \cite{ge2022miles} &    28 &  24.4 &  3.9 &  24 & 22.6 &  22.4 &  4.7 & 22.2          \\
     & VIOLET \cite{fu2021violet} &     24.8  &  23.9 & 4.5 &   26 & -  & -  & -  & -         \\
     & MVM \cite{fu2023empirical} & 24.8  &  23.9 & 4.5 &   26 & -  & -  & -  & -         \\
	\hline
 \multirow{5}{*}{CLIP \cite{radford2021learning}} & 
      CLIP4Clip \cite{luo2022clip4clip}&     42.6  &  25.4 & 6.7 & 37.7 & 41.4 &  19  & 7.9 &   38            \\
      & XCLIP \cite{ma2022x} &     43.2  &  39.8 &   8.7 & 41.5 & 45.6 &  41.2 &  10.3 & 40.2            \\
    & XPOOL \cite{gorti2022x} &    43.7  &  40.3 &  8.4 & 40.1 & 43.7 &  40.4 &  8.9 & 39.5         \\
     & EMCL \cite{jin2022expectation} &    46.3 &  40.2 &   6.9 & 41.7 & 44.8 &  42.1 &   9.3 & 42.3          \\
     & DiCoSA \cite{jin2023text} &     45.4  &  41.5 &  7.9 & 41.1 & 45.1 &   41.8 &  9.5 & 41.2           \\
	   \hline
		\end{tabular}}
		\caption{The table shows the results on DiDeMo \cite{anne2017localizing} dataset in both text-to-video and video-to-text retrieval settings. $Q$ denotes the performance (R@1 score) on the original unchanged dataset. $Q_{actrem}$, $Q_{objattrrem}$ and $Q_{synrem}$ is the R@1 score on datasets that have excluded actions, attributes and syntax respectively.}
        \label{tab: DIDEMO_rem}
	\end{center}
\end{table*}

\section{Experiments} \label{exper}
In this section, we explain the video retrieval models and datasets used for the proposed analysis.
\subsection{Models} \label{models}
We experiment with two categories of video retrieval models. The first category of models are pretrained on large scale video-text datasets like WebVid-2.5M \cite{bain2021frozen} and YT-Temporal-180M \cite{zellers2021merlot} and fine-tuned for downstream video retrieval datasets. These include Frozen-in-Time (FiT) \cite{bain2021frozen}, MCQ \cite{ge2022bridging}, MILES \cite{ge2022miles}, VIOLET \cite{fu2021violet} and MVM \cite{fu2023empirical}. The second category involves models that adapt pretrained image-text features such as CLIP \cite{radford2021learning} for the task of video retrieval. This category comprise of seven architectures namely TS2NET \cite{liu2022ts2}, CLIP4CLip \cite{luo2022clip4clip}, CLIP2Video \cite{fang2021clip2video}, XCLIP \cite{ma2022x}, XPOOL \cite{gorti2022x}, EMCL \cite{jin2022expectation} and DiCoSA \cite{jin2023text}.
\subsection{Datasets} \label{datasets}
We perform the evaluation on three video retrieval datasets: MSRVTT \cite{xu2016msr}, MSVD \cite{chen2011collecting} and DiDeMo \cite{anne2017localizing}. MSRVTT has 10000 videos and each video has multiple captions totalling 200K. We report the results on MSRVTT-9k split (9000 for training and 1000 for testing). MSVD consists of 1970 videos and 80K captions. The training split has 1300 videos and the test split has 670 videos. The captions in these datasets are single sentence. DiDeMo is made up of 10K videos and 40K descriptions. Following \cite{luo2022clip4clip}, we concatenate all the sentences and evaluate as paragraph-to-video retrieval. 
\subsection{Implementation} We use spacy\footnote{https://spacy.io/usage/linguistic-features} to identify parts-of-speech for all the words in the caption. We consider nouns, adverbs and adjectives as objects \& attributes, verbs as actions and rest of the parts-of-speech as syntax. We use the exact set up used by the state-of-the-art video retrieval models and measure the performance on all the augmented datasets.

\begin{comment}
\section{Methodology}
To correctly retrieve the ground truth video requires compositional understanding of several objects, attributes, entities and the hierarchical interactions among them in the text caption. However, currently there are no datasets to test the compositionality of video retrieval models. Therefore, we modify the original text captions by excluding, changing or adding certain parts. Each of the modified text caption is tailored to test the compositional knowledge of a particular important component of the original text caption essential in retrieving the correct ground truth video. The detailed list is elucidated  in the table ~\ref{tab:notation}. For easy understanding we explain the perturbations applied with an example below. Assume the original text query $Q$ as ``the squirrel ate the peanut out of the shell".
\end{comment} 
\begin{figure*}
     \centering
     \begin{subfigure}[b]{0.5\textwidth}
         \centering
         \includegraphics[width=\textwidth]{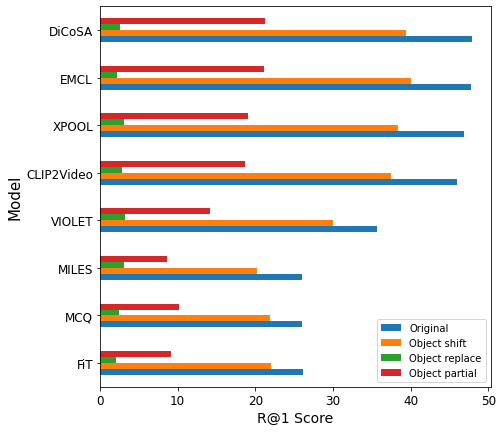}
         \caption{MSRVTT}
         \label{fig:msrvtt_objattabl}
     \end{subfigure}
     \hfill
     \begin{subfigure}[b]{0.49\textwidth}
         \centering
         \includegraphics[width=\textwidth]{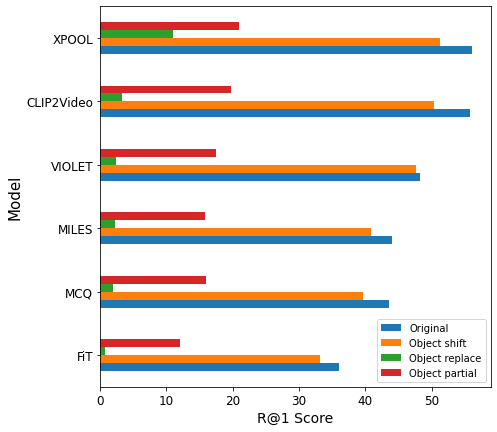}
         \caption{MSVD}
         \label{fig:msvd_objattabl}
     \end{subfigure}
     \caption{We perform ablation studies on the role of objects \& attributes in video retrieval. The video retrieval models are evaluated on three tasks namely: Object  shift ($Q_{objshift}$), Object  replacement ($Q_{objrep}$) and Object partial ($Q_{objpartial}$). Results show that swapping of objects has minor effect on performance followed by masking 50\% objects. The highest drop is seen when the objects are randomly replaced.  These ablation studies are performed on MSRVTT \cite{xu2016msr} and MSVD \cite{chen2011collecting} datasets}.
        \label{fig:objatrr_abl}
\end{figure*}

\begin{figure*}
     \centering
     \begin{subfigure}[b]{0.49\textwidth}
         \centering
         \includegraphics[width=\textwidth]{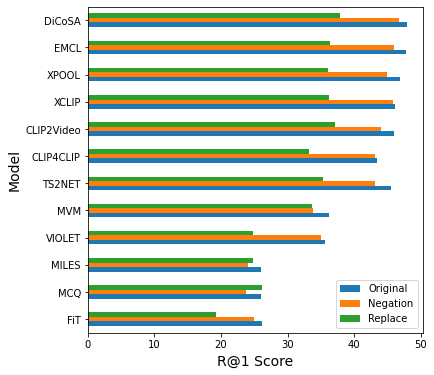}
         \caption{MSRVTT}
         \label{fig:msrvtt_actionabl}
     \end{subfigure}
     \hfill
     \begin{subfigure}[b]{0.49\textwidth}
         \centering
         \includegraphics[width=\textwidth]{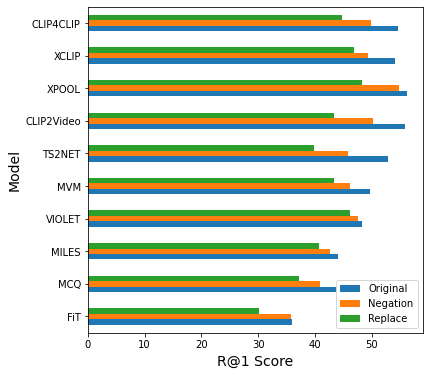}
         \caption{MSVD}
         \label{fig:msvd_actionabl}
     \end{subfigure}
     \caption{Figure shows the performance comparison (R@1 score) of video retrieval models on action ablation studies. The VR models are evaluated on captions with negated actions and replaced actions from MSRVTT \cite{xu2016msr} and MSVD \cite{chen2011collecting} datasets respectively. These studies illustrate that VR models have incomplete knowledge of negation and also are immune to action replacement in the captions}
        \label{fig:actionabl}
\end{figure*}

\begin{figure*}
     \centering
     \begin{subfigure}[b]{0.5\textwidth}
         \centering
         \includegraphics[width=\textwidth]{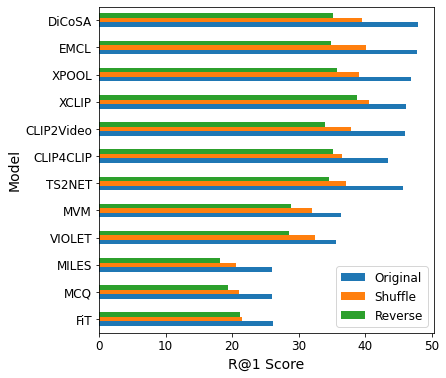}
         \caption{MSRVTT}
         \label{fig:msrvtt_order}
     \end{subfigure}
     \hfill
     \begin{subfigure}[b]{0.49\textwidth}
         \centering
         \includegraphics[width=\textwidth]{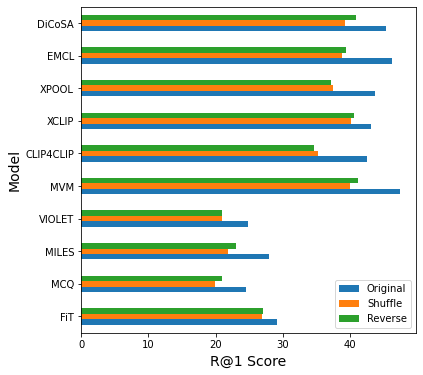}
         \caption{DiDeMo}
         \label{fig:didemo_order}
     \end{subfigure}
     \caption{Video retrieval performance (R@1) on word order task. We test the models on original (unchanged) captions, captions with shuffled word order and captions with reversed word order for MSRVTT and DiDeMo datasets. We demonstrate that VR models act like bag-of-words and do not require substantial word order information.}
        \label{fig:word_order}
\end{figure*}

\section{Results and Discussion} 
\subsection{Objects \& Attributes vs Actions vs Syntax: Do all of them matter?} \label{basic}
Our aim is to analyze the importance of three components: \textbf{\textcolor{Salmon}{objects \& attributes}}, \textbf{\textcolor{LimeGreen}{actions}} and \textbf{\textcolor{SkyBlue}{syntactics}} that make up a text query for retrieving videos. Hence, we test the video retrieval models with text captions that have missing objects \& attributes ($Q_{objattrrem}$), actions ($Q_{actrem}$) and syntax ($Q_{synrem}$).
Tables ~\ref{tab: MSRVTT_rem}, ~\ref{tab: MSVD_rem} and ~\ref{tab: DIDEMO_rem} show the results on MSRVTT, MSVD and DiDeMo datasets respectively. It is evident from the table that there is a drop in video retrieval performance when tested with text captions that don't have actions ($Q_{actrem}$). The drop is more pronounced among CLIP based models than pretrained video models. This shows that actions play a role for retrieving correct videos. 
However, we see that the performance drop is not as expected. Videos are time-series image frames which can have same attributes. In those scenarios, actions help in differentiating those videos. We see this effect when the R@1 is lower among pretrained video models and higher among CLIP based models. When the performance is lower, actions do not play a significant role in video retrieval and hence the videos can be retrieved without them in the text caption. On the contrary if R@1 score is higher, we see a notable decline. This is due to the robust video representations of CLIP based models as compared to pretrained video models. CLIP based models accurately encode video representations but, when the differentiating factor among videos i.e actions are missing in text captions leads to retrieval of incorrect videos. In short caption length datasets like MSRVTT and MSVD, we notice a significant drop in performance as compared with DiDeMo which is a paragraph ($>$ 1 sentences) dataset. This is because text captions in DiDeMo contains detailed description of the videos and hence, missing actions didn't lead to drop in performance as compared to MSRVTT and MSVD. It demonstrates that actions are not essential in paragraph-video retrieval.  

Next, we analyze the performance of video retrieval models tested with text captions without syntactics ($Q_{synrem}$). From the table, it is clear that there is a reduction in R@1 without the syntax in the text captions. It validates that syntactics are necessary for retrieving correct ground truth videos. For MSRVTT, we observe that models tested without syntax under-perform compared to actions in the text captions and the average difference in performance is 2\%. The reverse is true for MSVD and DiDeMo datasets where there is a huge difference of 9\%. In addition, we also notice that CLIP based models are more sensitive to syntax than pretrained video models. Finally, we evaluate the video retrieval models on text captions in the absence of objects \& attributes ($Q_{objattrrem}$). As seen from the results, these models perform poorly (a drop in 20\%) which underscores the significance of objects \& attributes. We also notice that $Q_{attrrem}$ trails $Q_{actrem}$ and $Q_{synrem}$ by a huge margin. This difference is more striking among CLIP based models as opposed to pretrained video models. 

%To summarize from the results, objects \& attributes are the most critical components followed by actions and semantics. 

\subsection{What role do Objects \& Attributes play in video retrieval?} \label{objabl}
In the previous sections (\S\ref{basic}), findings from our experimental results suggested that objects \& attributes are the most important component in text captions while retrieving videos. To investigate further, we perform additional detailed studies on their importance. In captions there can be multiple objects \& attributes and every pair of object \& attribute is distinct from the other. Any slight modification in the pairs can totally change their correspondence and thereby the ground truth video and hence, video retrieval models should be able to account for these changes. We perform a test in which we interchange the places of objects while keeping rest of the caption same $Q_{objshift}$. In the second study, we randomly replace objects in the caption $Q_{objrep}$ and evaluate the models on the modified ones. The final ablation involves keeping just half the objects in the captions ($Q_{objpartial}$). This is to assess if VR models adapt any shortcuts and still retrieve correct videos without the critical information. Figure ~\ref{fig:objatrr_abl} demonstrates the results for these studies. As shown in the figure, there is a slight deterioration of video retrieval performance when there is a object shift in the caption. The drop is a meagre 5.5\% for MSRVTT and 3.6\% in case of MSVD dataset. It demonstrates that VR models do not quite fully understand the relationship between object and its attribute. On the other hand if the objects are randomly replaced ($Q_{objrep}$) with different unrelated objects, there is a massive degradation in R@1 score. In fact, the performance is quite similar to the models performance tested on captions without objects \& attributes. These results prove that video retrieval models are extremely sensitive to alteration of objects. Figure ~\ref{fig:objatrr_abl} shows that there is a noticeable fall in performance when the VR models have access to just 50\% of the object data in the captions. The R@1 score lags by 30\% in MSRVTT and 22\% in MSVD datasets.
It reinforces the aforesaid extreme sensitivity nature of the retrieval models towards objects. Furthermore, random object replacement performs far worse than partial objects in the captions. This highlights that factual object description even though 50\% is much more crucial than access to the entire caption albeit with incorrect objects.

\subsection{Do VR models pay attention to actions?} \label{actionabl}
We demonstrated in the section ~\ref{basic} that actions play a role in video retrieval. Now, this raises an important question \textit{\textbf{How much attention do VR pay to actions in the captions?}} To investigate this we perform certain ablation studies on the action understanding of VR models in the text captions. We replace the action word with the negation of it ($Q_{actneg}$) and test the performance of VR models on the newly formed captions. In parallel, the actions in the captions are randomly replaced with different actions ($Q_{actrep}$) and VR models are evaluated on the altered captions. In Figure ~\ref{fig:actionabl}, we provide the results of VR models tested on captions with negated ($Q_{actneg}$) and replaced ($Q_{actrep}$) actions. From the figure, it is evident that action negation ($Q_{actneg}$) achieves comparable results to $Q$ and there is a slight drop in performance in case of action replacement ($Q_{actrep}$). Most of the actions are expressed in positive sense in these datasets and this is not always the case. For a fine-grained description of videos, the actions of the static objects can be communicated in a negation form. So, naturally video retrieval models are expected to understand the negation in captions. However, we notice that action negation has similar performance as original captions which demonstrates that VR models lack the capability of action negation sense. Next, we randomly replace the actions with a different action and test the attention of VR models. In an ideal scenario, the performance of these models should drop drastically as the replaced actions do not correspond to that of the ones in ground truth videos. Nevertheless, we see that the R@1 score of action replacement ($Q_{actrep}$) is only slightly less than original caption $Q$. In fact the average drop in R@1 is only 6.8\% in MSRVTT and 7.5\% in MSVD. Hence even though the actions are important in video retrieval, VR models use other influential information such as objects \& attributes to retrieve ground truth videos. 

\subsection{Does word order of text captions matter?} \label{wordorder}
In the figures \ref{fig:msrvtt_order} and \ref{fig:didemo_order}, we present the findings on the word order evaluation. First we observe that models tested on datasets without word order perform worse than the original dataset. The R@1 score is reduced on average by 6.3\%  and 9.1\% on shuffled ($Q_{s}$) and reversed ($Q_{r}$) MSRVTT captions respectively. On a similar note the performance drops 5.5\% on shuffled and 5\% on reversed DiDeMo dataset. Additionally, the R@1 decrease is more pronounced on reversed captions than shuffled. This is surprising as the object-action order is preserved in reversed captions in contrast with shuffled.
This shows that models adapt bag-of-words approach for syntactic understanding of captions and positioning of object-action order doesn't matter. A possible explanation for this behaviour is: all the video retrieval models use pretrained language models as their text encoder. Recent studies have shown that \cite{sinha2021masked,madasu-srivastava-2022-large} distributional information is preserved even though the syntactic word order is disturbed and hence, LMs leverage it for hierarchical text understanding. Surprisingly, video retrieval models manifest the same behaviour in caption understanding.

\section{Conclusion}
In this work, we proposed a comprehensive investigation of compositional and syntactic understanding of video retrieval models. For this study we put forward 10 different tasks to evaluate models reasoning of objects \& attributes, actions and syntax for retrieving videos. We experiment with a wide range of 12 state-of-the-art video retrieval models and 3 standard benchmarks. We show that video retrieval performance is heavily impacted by objects \& attributes and lightly by syntactics. Furthermore, our results also reveal that word order matter less for video retrieval models. These results shed an important light on the inner workings of video retrieval models. We believe the future works can utilize these findings to design compositional aware video retrieval models. 
{
    \small
    \bibliographystyle{ieeenat_fullname}
    \bibliography{main}
}


\end{document}